\documentclass{opt2022} 
\usepackage{hyperref}
\usepackage{url}
\usepackage{multirow}

\usepackage{ulem}

\usepackage{graphicx}
\usepackage{caption}
\usepackage{graphics}

\newcommand{\R}{\mathbb{R}}

\newcommand{\p}{{\rm I}\kern-0.18em{\rm P}}
\newcommand{\E}{{\rm I}\kern-0.18em{\rm E}}

\title[Improved Deep Neural Network Generalization Using mSAM]{Improved Deep Neural Network Generalization Using m-Sharpness-Aware Minimization}

\optauthor{%
\Name{Kayhan Behdin}\thanks{This work was done when Kayhan Behdin was an intern at LinkedIn during summer 2022.} \Email{behdin1675@gmail.com}\\
\addr LinkedIn Corporation, USA\\
(Massachusetts Institute of Technology, USA)
\AND
\Name{Qingquan Song} \Email{qsong@linkedin.com}\\
\Name{Aman Gupta} \Email{amagupta@linkedin.com}\\
\Name{David Durfee} \Email{ddurfee@linkedin.com}\\
\Name{Ayan Acharya} \Email{ayacharya@linkedin.com}\\
\Name{Sathiya Keerthi} \Email{keselvaraj@linkedin.com}\\
\addr LinkedIn Corporation, USA
\AND
\Name{Rahul Mazumder}\thanks{This work was done when 
Rahul Mazumder was a consultant for LinkedIn.}
\Email{rmazumder@linkedin.com}\\
\addr LinkedIn Corporation, USA\\
(Massachusetts Institute of Technology, USA) }

\begin{document}

\maketitle

\begin{abstract}
Modern deep learning models are over-parameterized, where the optimization setup strongly affects the generalization performance. A key element of reliable optimization for these systems is the modification of the loss function. Sharpness-Aware Minimization (SAM) modifies the underlying loss function to guide descent methods towards flatter minima, which arguably have better generalization abilities. In this paper, we focus on a variant of SAM known as mSAM, which, during training, averages the updates generated by adversarial perturbations across several disjoint shards of a mini-batch. Recent work suggests that mSAM can outperform SAM in terms of test accuracy. However, a comprehensive empirical study of mSAM is missing from the literature---previous results have mostly been limited to specific architectures and datasets. To that end,  this paper presents a thorough empirical evaluation of mSAM on various tasks and datasets. We provide a flexible implementation of mSAM and compare the generalization performance of mSAM to the performance of SAM and vanilla training on different image classification and natural language processing tasks. We also conduct careful experiments to understand the computational cost of training with mSAM, its sensitivity to hyperparameters and its correlation with the flatness of the loss landscape. Our analysis reveals that mSAM yields superior generalization performance and flatter minima, compared to SAM, across a wide range of tasks without significantly increasing computational costs.
\end{abstract}


\section{Introduction}

In recent years, overparameterized deep neural networks (DNNs) have become a staple of modern machine learning advancements and crucial to many domains. In particular, overparameterization has been successful in furthering the state-of-the-art performance for domains like image understanding~\cite{alexnet, resnets, tan2019efficientnet}, natural language processing (NLP)~\cite{vaswani2017attention, devlin2018bert, liu2019roberta} and recommender systems~\cite{guo2017deepfm, naumov2019deep}. 

Training DNNs requires minimizing complex and non-convex loss functions with an abundance of local minima. Interestingly, different local minima can have varying loss values and generalizability on unseen data. Thus, it is essential to carefully choose an optimization scheme that can seek out minima that yield strong generalization performance. In recent years, a wide range of optimization algorithms has been developed for various domains, e.g. stochastic gradient descent (SGD), heavy-ball momentum~\cite{sutskever2013importance}, Adam~\cite{kingma2014adam}, LAMB~\cite{you2019large} among others. These methods help yield strong generalization performance when coupled with suitable regularization techniques. 

Recently, extensive work has been done to study the correlation between the geometry of the loss landscape and generalization performance~\cite{flatness2, flatness1, xiss21, wuws22, hawl21, saqu18}. The newly proposed Sharpness-Aware Minimization (SAM) algorithm~\cite{sampaper} leverages this correlation between the flatness of minima and generalization performance by seeking to obtain flatter solutions during training, leading to superior generalization for a panoply of tasks and domains. In this paper, we focus on a variant of SAM called mSAM~\cite{sampaper}. 
Instead of using a single adversarial perturbation for the entire mini-batch
mSAM executes SAM-like training by averaging the updates generated by adversarial perturbations across several disjoint shards of a mini-batch. The name mSAM stems from the usage of $m$ such disjoint shards. SAM represents flatness by looking at what happens in one particular adversarial direction, whereas mSAM leverages several such directions; which may better represent flatness.


\paragraph{Related Work:}

Although the sharpness of the loss landscape can be calculated using several different measures (for example, the largest eigenvalue~\cite{sharpnesspaper} or trace~\cite{minimumsharpness} of the Hessian of the loss), most of these measures are computationally too expensive for practical purposes. The main idea behind the SAM~\cite{sampaper} algorithm is to encourage the network to seek regions where the worst loss value in a local neighbourhood is not too large (see Section~\ref{sec:algs} for more details). This proxy of sharpness lends itself to easy computation, unlike the measures of sharpness described earlier. SAM has sparked interest in sharpness-aware training, resulting in several variants~\cite{gsampaper, looksam, samforfree}.

In this paper, we focus on a less-understood variant of SAM, known as mSAM. While SAM uses a single adversarial perturbation for a mini-batch, mSAM divides the mini-batch into disjoint shards and computes updates using adversarial perturbations applied to each shard. The updates are then averaged, resulting in a single mSAM update. In~\cite{sampaper}, mSAM is used \emph{implicitly} to reduce the computational cost of SAM by avoiding synchronization across multiple GPUs (referred to as ``accelerators'' hereinafter). Thus, each accelerator's shard of data ends up with its own adversarial perturbation. Recently, it has been observed via limited experimentation that mSAM results in better generalization performance~\cite{sampaper,samvit,nonsensemsam}. Although the authors of \cite{icmlPaper} present mathematical expressions for mSAM, their analysis is primarily focused on a particular version of mSAM (see Section~\ref{sec:algs} for more details). The experiments are also limited to image classification tasks on small architectures. As a result, the current understanding of mSAM is limited, and a comprehensive empirical study of its performance on state-of-the-art architectures and datasets is missing from the literature.


\paragraph{Our Contributions:}
We conduct a thorough empirical study of the mSAM algorithm and perform the following explorations that extol the value of mSAM.
\textbf{(i)} Starting from the mathematical description of mSAM, we present an \emph{explicit} flexible implementation of mSAM that does not rely on accelerator synchronization and is compatible with any single/multi-accelerator setup. \textbf{(ii)} We conduct extensive experiments on a wide variety of tasks using images and NLP data, leveraging architectures like Convolutional Neural Networks (CNNs) and transformers. In our experiments, mSAM consistently outperforms SAM and vanilla training. \textbf{(iii)} We also conduct careful experiments to understand the computational cost of training with mSAM, its sensitivity to hyperparameters like $m$ and its correlation with the flatness of the loss landscape. Our thorough empirical investigation reveals that mSAM substantially improves the generalization performance of SAM without significantly increasing the computational cost of training.


\section{Algorithm}\label{sec:algs}

mSAM is based on the SAM algorithm that aims to obtain flat solutions to the empirical loss function. In particular, SAM tries to find a solution that minimizes the worst-case loss in a ball around the solution. Mathematically, let $\mathcal{S}=\{(x_i,y_i), i\in[n]: x_i\in\mathcal{X},y_i\in\mathcal{Y}\}$ be a dataset of $n$ samples, where $\mathcal{X}$ is the set of features and $\mathcal{Y}$ is the set of outcomes. Moreover, let $\ell:\R^d\times \mathcal{X}\times \mathcal{Y}\mapsto \R$ be a differentiable loss function, where $d$ is the number of model parameters. The empirical loss over the dataset $\mathcal{S}$ is defined as $ \mathcal{L}_{\mathcal{S}}(w)=\sum_{i=1}^n \ell(w;x_i,y_i)/n$, where $w$ parameterizes the neural network. With this notation in place, the SAM loss function is defined as~\cite{sampaper}:
\begin{equation}\label{sam-loss}
    \mathcal{L}^{SAM}_{\mathcal{S}}(w)=\max_{\|\epsilon\|_p\leq \rho}\mathcal{L}_{\mathcal{S}}(w+\epsilon)
\end{equation}
for some $p\geq 1$. In this work, we use $p=2$. In practice, however, the maximization step in~\eqref{sam-loss} cannot be done in closed-form. Hence, authors in \cite{sampaper} use a first-order approximation to $\mathcal{L}_S$ to simplify~\eqref{sam-loss} as
\begin{equation}\label{sam-loss-linear}
  \mathcal{L}^{SAM}_{\mathcal{S}}(w)\approx\max_{\|\epsilon\|_2\leq \rho}\mathcal{L}_{\mathcal{S}}(w)+\epsilon^{\top} \nabla\mathcal{L}_S(w).  
\end{equation}
It is easy to see that the maximum in Problem~\eqref{sam-loss-linear} is achieved for
\begin{equation}
    \hat{\epsilon}=\rho \nabla\mathcal{L}_{\mathcal{S}}(w)/\|\nabla\mathcal{L}_{\mathcal{S}}(w)\|_2.
\end{equation}
As a result, $\mathcal{L}_{\mathcal{S}}^{SAM}\approx \mathcal{L}_{\mathcal{S}}(w+\hat{\epsilon})$. This leads to the gradient  
\begin{equation}\label{sam-grad-2nd}
    \nabla\mathcal{L}_{\mathcal{S}}^{SAM}(w)\approx\nabla_{w}\left[\mathcal{L}_{\mathcal{S}}(w+\hat{\epsilon})\right]=\frac{\partial(w+\hat{\epsilon})}{\partial w}\nabla\mathcal{L}_{\mathcal{S}}(w+\hat{\epsilon}).
\end{equation}
However, calculating $\partial(w+\hat{\epsilon})/\partial w$ in~\eqref{sam-grad-2nd} involves second order terms that require access to Hessian, which can be computationally inefficient in practice. Thus, by ignoring second order terms in~\eqref{sam-grad-2nd}, gradient of the SAM loss, is approximated as~\cite{sampaper}:
\begin{equation}\label{sam-gradient}
    \nabla \mathcal{L}^{SAM}_{\mathcal{S}}(w) \approx \nabla \mathcal{L}_{\mathcal{S}}(w+\rho \nabla\mathcal{L}_{\mathcal{S}}(w)/\|\nabla\mathcal{L}_{\mathcal{S}}(w)\|_2)
\end{equation}
which is used in the SAM algorithm (for example, in conjunction with SGD). We refer to~\cite{sampaper} for more details and intuitions about SAM. We call the inner gradient calculations on the right-hand side of~\eqref{sam-gradient} as the SAM ascent step and the outer gradient calculations as the gradient step. mSAM~\cite{sampaper} is a variation of the SAM algorithm. In general, for mSAM, a minibatch of data  $\mathcal{S}$ is further divided into $m$ smaller disjoint shards (aka ``micro-batches''), such as $\mathcal{S}_1,\cdots,\mathcal{S}_m$ where $\cup_{i=1}^m \mathcal{S}_i=\mathcal{S}$. For simplicity, we assume $|\mathcal{S}_1|=\cdots=|\mathcal{S}_m|=|\mathcal{S}|/m$ although such an assumption is not necessary in general. The mSAM loss is a variation of the SAM loss, defined as:
\begin{equation}\label{msam-loss}
     \mathcal{L}^{mSAM}_{\mathcal{S}}(w) = \frac{1}{m}\sum_{i=1}^m \max_{\|\epsilon^{(i)}\|_2\leq \rho} \mathcal{L}_{\mathcal{S}_i}(w+\epsilon^{(i)}).
\end{equation}
Intuitively, mSAM is a version of SAM where the ascent step (or the weight perturbation) of SAM is done independently on each micro-batch using different $\epsilon^{(i)}$, instead of using an average perturbation such as ${\epsilon}$ for all micro-batches. The mSAM gradient can thereby be derived as:
\begin{equation}\label{msam-gradient}
        \nabla \mathcal{L}^{mSAM}_{\mathcal{S}}(w) = \frac{1}{m}\sum_{i=1}^m \nabla \mathcal{L}_{\mathcal{S}_i}(w+\rho \nabla\mathcal{L}_{\mathcal{S}_i}(w)/\|\nabla\mathcal{L}_{\mathcal{S}_i}(w)\|_2),
\end{equation}
where~\eqref{msam-gradient} is a first-order approximation to the gradient of~\eqref{msam-loss}. We also note that the loss~\eqref{msam-loss} is  related to the mSAM definition of~\cite{icmlPaper}.  See Table~\ref{sam-table} for a side-by-side comparison of SAM and mSAM, and their different implementations.

\begin{table}[t]
\small
\renewcommand{\arraystretch}{1.2}
\caption{Comparison of SAM with Different mSAM Implementations}
\label{sam-table}
\vspace{-0.4cm}
\begin{center}
\begin{tabular}{c|c|ccc}
 & SAM &  \multicolumn{3}{c}{mSAM} \\
 \hline
 Loss function & $\max_{\|\epsilon\|_2\leq \rho} \sum_{i=1}^m  \mathcal{L}_{\mathcal{S}_i}(w+\epsilon)/m$ & \multicolumn{3}{c}{$\sum_{i=1}^m \max_{\|\epsilon^{(i)}\|_2\leq \rho} \mathcal{L}_{\mathcal{S}_i}(w+\epsilon^{(i)})/m$}\\
 Ascent step & $\hat{\epsilon}\propto \rho {\sum_{i=1}^m \nabla\mathcal{L}_{\mathcal{S}_i}(w)}/m$ &  \multicolumn{3}{c}{$\hat{\epsilon}^{(i)}\propto\rho{ \nabla\mathcal{L}_{\mathcal{S}_i}(w)},~i\in[m]$} \\
  Gradient & $g= \sum_{i=1}^m\nabla \mathcal{L}_{\mathcal{S}_i}({w}+\hat{\epsilon})/m$ &  \multicolumn{3}{c}{$g= \sum_{i=1}^m \nabla \mathcal{L}_{\mathcal{S}_i}(w+\hat{\epsilon}^{(i)})/m$} \\
 Implementations & \cite{sampaper} & \cite{sampaper} & \cite{icmlPaper} & Ours \\
Possible $m$ values & - & $\#$ of accelerators & flexible & flexible \\
Processor support & Multiple & Multiple & Single & Multiple
\end{tabular}
\end{center}
\end{table}

An important distinction between our work and prior work is that we treat $m$ as a model hyper-parameter to improve generalization. In particular, in mSAM implementation of~\cite{sampaper}, the value of $m$ is fixed to the number of hardware accelerators, micro-batch $i$ is the part of the data that is loaded onto accelerator $i$, and each accelerator uses a separate perturbation, simulating the effect of mSAM. With this implementation, $m$ is an artefact of the hardware setup. On the other hand, the analysis of~\cite{icmlPaper} mostly concerns the value $m=|\mathcal{S}|$. In contrast, we consider a wide range of values for $m$ in our experiments---this offers the flexibility to choose an appropriate value of $m$ that leads to a better generalization performance. Moreover, our implementation supports any single/multi-accelerator setup and allows the user to set an appropriate value of $m$.




\section{Numerical Experiments}
This section compares mSAM to SAM and vanilla optimization methods (i.e. without sharpness-aware modification) on various model architectures and datasets. We report the average and standard deviation of accuracy on the test data over five independent runs. We also note that we use the same values of hyper-parameters for all algorithms, where $\rho$ is chosen based on the best validation performance for SAM and other hyper-parameters are chosen based on the best validation error for vanilla methods. Moreover, although it is possible to use different values of $\rho$ for each micro-batch in mSAM, doing so requires tuning numerous hyper-parameters (as we usually take $m$ to be large) which is computationally infeasible. Therefore, we use the same value of $\rho$ for all micro-batches.

\subsection{Image Classification}\label{imagesec}
In our first set of experiments on image classification datasets,
we compare the performance of mSAM, SAM and vanilla methods with multiple CNN architectures such as ResNets~\citep{resnets} and WideResNet~\citep{wideresnet}. We use CIFAR10/100~\citep{cifar}
datasets as our test bed.

The average accuracies for the experiments with CIFAR data are reported in Table~\ref{cifar-table}. These experiments are done on four NVidia V100 GPUs, with an effective batch size of $512$ and the value of $m=32$ for mSAM. Other hyper-parameters used to produce these results can be found in Appendix~\ref{imageparam}. Overall, mSAM leads to better accuracy than SAM and vanilla methods in all cases reported in Table~\ref{cifar-table}. In particular, mSAM achieves the best improvement in CIFAR100 data (about $2\%$ on ResNet50 and about $1\%$ on WRN-28-10) which is a more challenging dataset compared to CIFAR10.
\begin{table}[t]
\small
\caption{Accuracy Results for CNN Architectures}
\label{cifar-table}
\begin{center}
\begin{tabular}{ccccc}
Dataset & Model & Vanilla & SAM &  mSAM 
\\ \hline 
\multirow{3}{*}{CIFAR 10} & ResNet50      & $95.45\pm 0.10$  & $96.09\pm0.11$ &  $96.40\pm0.06$\\ 
& WRN-28-10     & $95.92\pm0.12$ & $96.90\pm0.05$ &  $96.95\pm0.04$ \\
& ViT-B/16     & $97.68\pm0.03$  & $97.75\pm0.06$ &  $98.29\pm0.08$ \\
\multirow{3}{*}{CIFAR 100} & ResNet50      & $80.68\pm0.13$ & $81.49\pm0.18$ & $83.37\pm0.10$\\
& WRN-28-10     &$81.01\pm 0.19$  & $82.93\pm0.13$ &  $84.07\pm0.06$\\
 & ViT-B/16      & $88.02\pm0.17$ & $88.75\pm0.07$ &   $89.00\pm 0.17$\\

\end{tabular}
\end{center}
\end{table}

Following the recent results that suggest that sharpness-aware optimization can improve the training quality of ViTs~\citep{samvit}, we conduct additional experiments on a ViT architecture. In particular, we use the pre-trained ViT-B/16 checkpoint from~\citep{vitpaper} and fine-tune the model on CIFAR10/100 data independently. The average accuracy results for ViT fine-tuning are reported in Table~\ref{cifar-table}. Similar to the CNN case, mSAM leads to better results than SAM that improves upon vanilla optimization further. We note that in these experiments, $m=32$ is fixed for mSAM.



\subsection{NLP Fine-tuning}
Our next set of experiments is based on NLP analysis. We select four tasks from GLUE benchmark~\citep{glue}. In particular, we choose COLA and MRPC as two small datasets and SST-2 and QQP as two larger datasets for empirical evaluation. The fine-tuning experiments with the RoBERTa-base model~\citep{roberta} are performed on four NVidia V100 GPUs with an effective batch size of $32$. For the ease of reproduction of the results, we tabulate all the hyper-parameters used in Appendix~\ref{glueparam}. For the fine-tuning experiments, we report the average value of Matthews Correlation Coefficient for COLA, and average accuracy for other datasets in Table~\ref{nlp-table}.
Overall, mSAM performs better than the baseline methods on these datasets. However, the variance among different runs is comparably high for smaller datasets such as COLA and MRPC. On the other hand, the results on larger data such as SST-2 and QQP are expectedly more robust across different runs.


\section{A Deeper Investigation of mSAM}
To further understand the mSAM algorithm, we design and report some experiments in this section. Additional experimental results are moved to the Appendix~\ref{app:switch}.

\begin{figure*}[t!]
     \centering
\begin{tabular}{cc}
 ResNet50 & WRN-28-10 \\
     \includegraphics[scale=0.40]{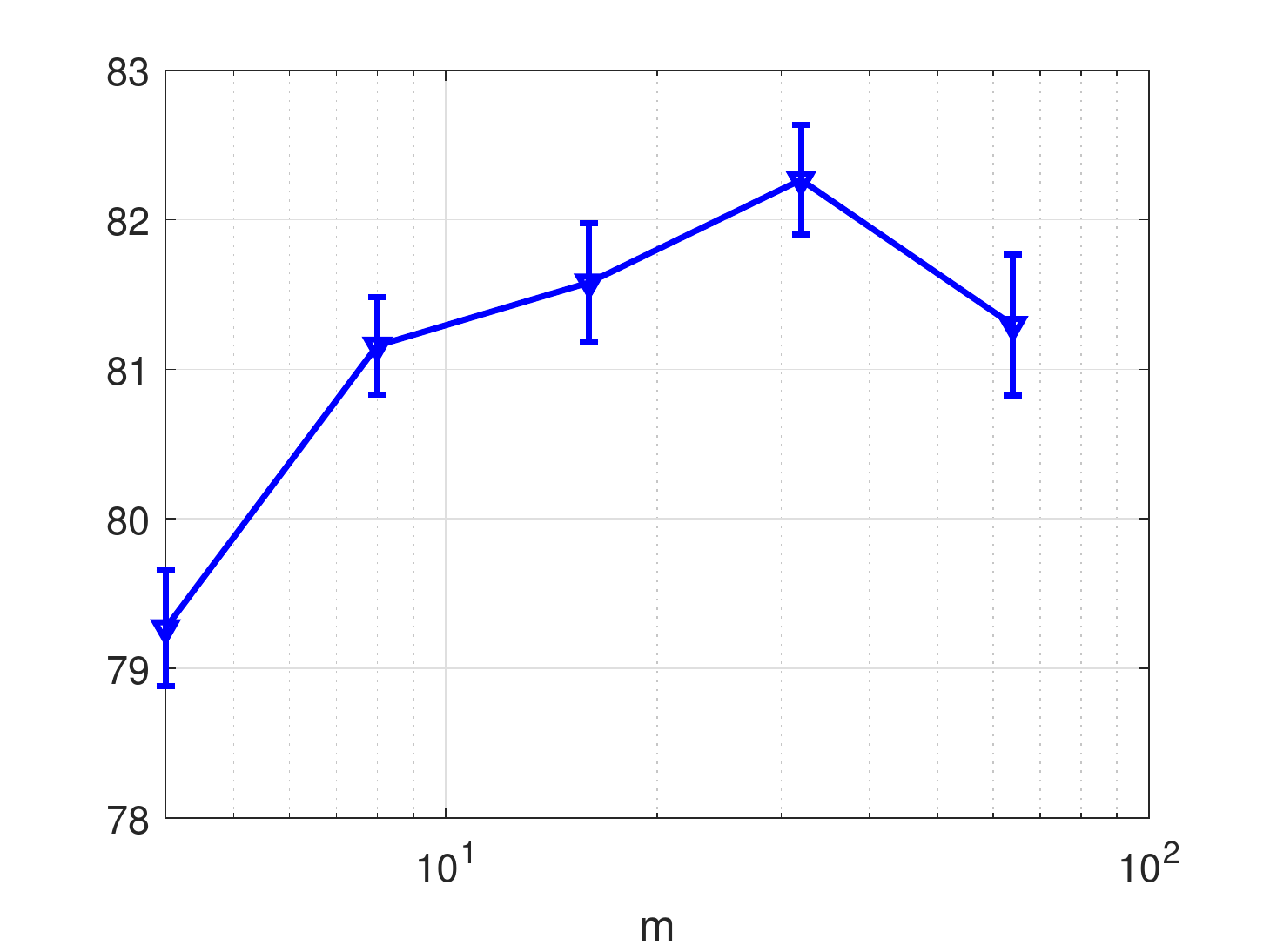}&   
     \includegraphics[scale=0.40]{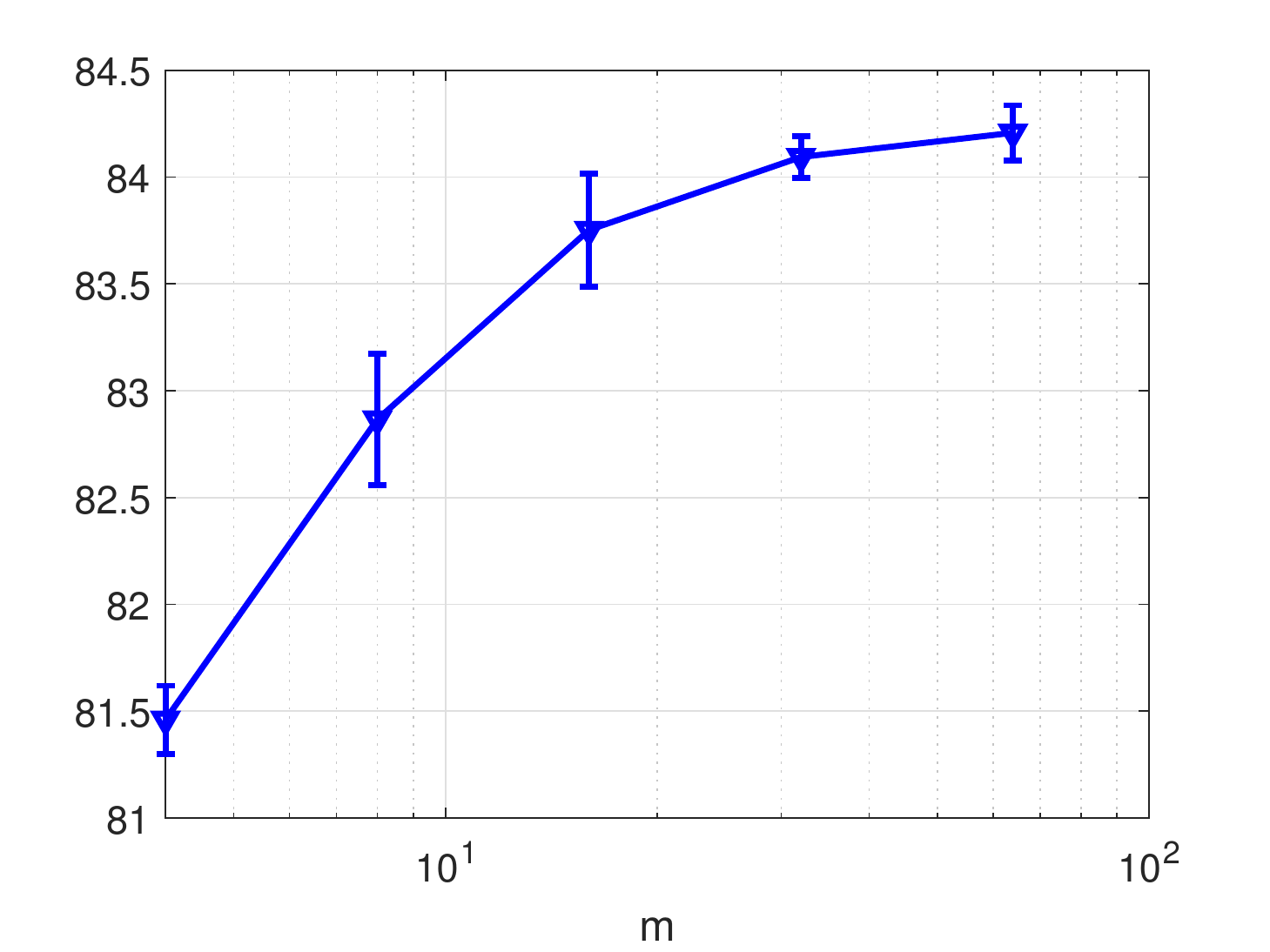}  
\end{tabular}
\vspace{-0.4cm}
        \caption{Effect of varying $m$}
        \label{fig:mb}
\end{figure*}

\paragraph{Effect of varying $m$:}
In our experiments, we have observed that a larger value of $m$ often leads to better test accuracy. We recover SAM by setting $m=1$, which produces inferior results. To test this hypothesis, we set up the experiments with the CIFAR100 dataset on two CNNs, ResNet50 and WRN-28-10 in the same setup as in Section~\ref{imagesec}. We run mSAM for different values of $m\in\{4,8,16,32,64\}$. The accuracy results for these experiments are shown in Figure~\ref{fig:mb}.

Increasing $m$ improves the performance up to $m\approx 32$. However, a value of $m$ larger than this threshold either leads to worse performance or marginal improvements, so increasing $m$ does not necessarily result in better generalization. Intuitively speaking, when the micro-batch is too small, the perturbation derived according to the micro-batch might not be a good estimate of the actual SAM perturbation, leading to worse performance. We leave the theoretical analysis of such a phenomenon an interesting direction for future research. We also note that understanding how the optimal value of $m$ and batch size interact is an open question left for future work.

\begin{table}[t!]
\small

\vspace{0.3cm}
\centering
\begin{minipage}{0.5\linewidth}
\centering
\caption{Accuracy Results for GLUE Tasks}
\label{nlp-table}

\begin{tabular}{cccc}
Task &Vanilla & SAM & mSAM ($m=8$)
\\ \hline 
COLA        & $63.66\pm2.46$ & $64.30\pm0.49$ & $64.57\pm 0.66$\\
MRPC        & $89.79\pm0.05$ & $90.37\pm0.13$ & $90.92\pm0.16$ \\
SST-2        & $94.27\pm0.18$ & $95.21\pm0.12$  & $95.38\pm0.10$  \\
QQP        & $91.70\pm0.11$ & $92.13\pm0.02$ &  $92.18\pm0.03$ \\
\end{tabular}

\end{minipage}
\hfill
\begin{minipage}{0.41\linewidth}

\centering
    \caption{$\lambda_{\max}$ for CNNs}
\label{eigen-table}
\begin{center}
\begin{tabular}{cccc}
 Model & Vanilla & SAM &  mSAM
\\  
\hline 
ResNet50     &  $26\pm2$  & $21\pm3$  & $18\pm1$ \\
WRN-28-10     &   $92\pm4$ & $30\pm2$  & $17\pm1$ \\
\end{tabular}
\end{center}

\end{minipage}

\end{table}

\paragraph{Are mSAM solutions flat?}
The SAM algorithm hypothesizes that flat solutions generalize better. Since mSAM consistently outperforms SAM, it is worth investigating if mSAM settles for even flatter solutions. To that end and to quantify sharpness, we calculate the largest eigenvalue of the Hessian of the loss function $\mathcal{L}_{\mathcal{S}}(w)$ at the final solution, denoted as $\lambda_{\max}$. This metric is widely accepted to be a good indicator of the sharpness/flatness of a solution~\cite{sharpnesspaper}. We calculate $\lambda_{\max}$ over the full train data using the power iteration, as implemented by~\cite{hessian-eigenthings}. We use the ResNet50 and WRN-28-10 models trained on CIFAR100 (see Section~\ref{imagesec}) to calculate $\lambda_{\max}$. The average results for these experiments are reported in Table~\ref{eigen-table}. We see that mSAM leads to solutions with smaller $\lambda_{\max}$ than SAM and vanilla SGD, confirming our conjecture.

\paragraph{mSAM runtime:}\label{app:runtime}
A general misconception about mSAM is that it is computationally inefficient, as the total number of forward-backward passes in the network is multiplied by $m$ \cite{nonsensemsam}. However, note that these passes are performed on micro-batches, which are $m$ times smaller than the actual minibatch. Hence, the overall computational cost gets amortized and is never as high as $m$ times the cost of SAM. In practice, on large networks, the runtime of mSAM is \emph{only} $1.2$-$1.3$ times more compared to SAM. In Appendix~\ref{app:switch}, we present numerical evidence for mSAM efficiency and discuss a few hybrid algorithms to reduce the computational cost of mSAM even further.


\section{Discussion}
The empirical study presented in this paper shows that mSAM outperforms SAM on different datasets (image classification and NLP tasks) and model architectures (CNNs and transformers); the exact performance gap is primarily a function of the data and the architecture. Moreover, our experiments suggest that mSAM does not necessarily incur a substantially higher computational cost than SAM, making it amenable for large-scale problems. An exciting avenue of work is the theoretical justification of better generalization abilities of mSAM. Our analysis suggests that mSAM leads to flatter solutions than SAM, which may explain the better generalization performance. However, the formal proof regarding why mSAM promotes a flatter minimum is left for future work.

\section{Acknowledgements}
Kayhan Behdin contributed to this work while he was an intern at LinkedIn during summer 2022. This work is not a part of his MIT research. Rahul Mazumder contributed to this work while he was a consultant for LinkedIn (in compliance with MIT’s outside professional activities policies). This work is not a part of his MIT research.


\newpage
\bibliography{ref}

\clearpage
\newpage
\appendix
\numberwithin{table}{section}
\numberwithin{figure}{section}


\section{mSAM and Computational Efficiency}\label{app:switch}

Since SAM requires two forward-backwards passes for each batch of data, SAM is almost twice as slow as vanilla training. In our experiments, mSAM appears to be slower than SAM, although not $m$ times slower, as suggested by, for example~\cite{nonsensemsam}. To demonstrate that, we report the average runtime of our experiments from Section~\ref{imagesec} on CIFAR100 data in Table~\ref{runtime-table}. Expectedly, SAM is twice as slow as the vanilla method. Interestingly, in the worst case, mSAM is \emph{only} twice as slow as SAM, and in the best case, the computational penalty is only about $10\%$ of SAM. The primary reason is that the overall computation complexity of forward-backwards passes for mSAM and SAM is not too different. In each epoch, mSAM performs $m$ times more passes on micro-batches that are $m$ times smaller. Encouragingly, mSAM appears more efficient on larger architectures such as ViT-B/16 and WRN-28-10. 

\begin{figure}[t]
\centering
\caption{Effect of switching training algorithm}
\begin{tabular}{cc}
 ResNet50 & WRN-28-10 \\
     \includegraphics[scale=0.50]{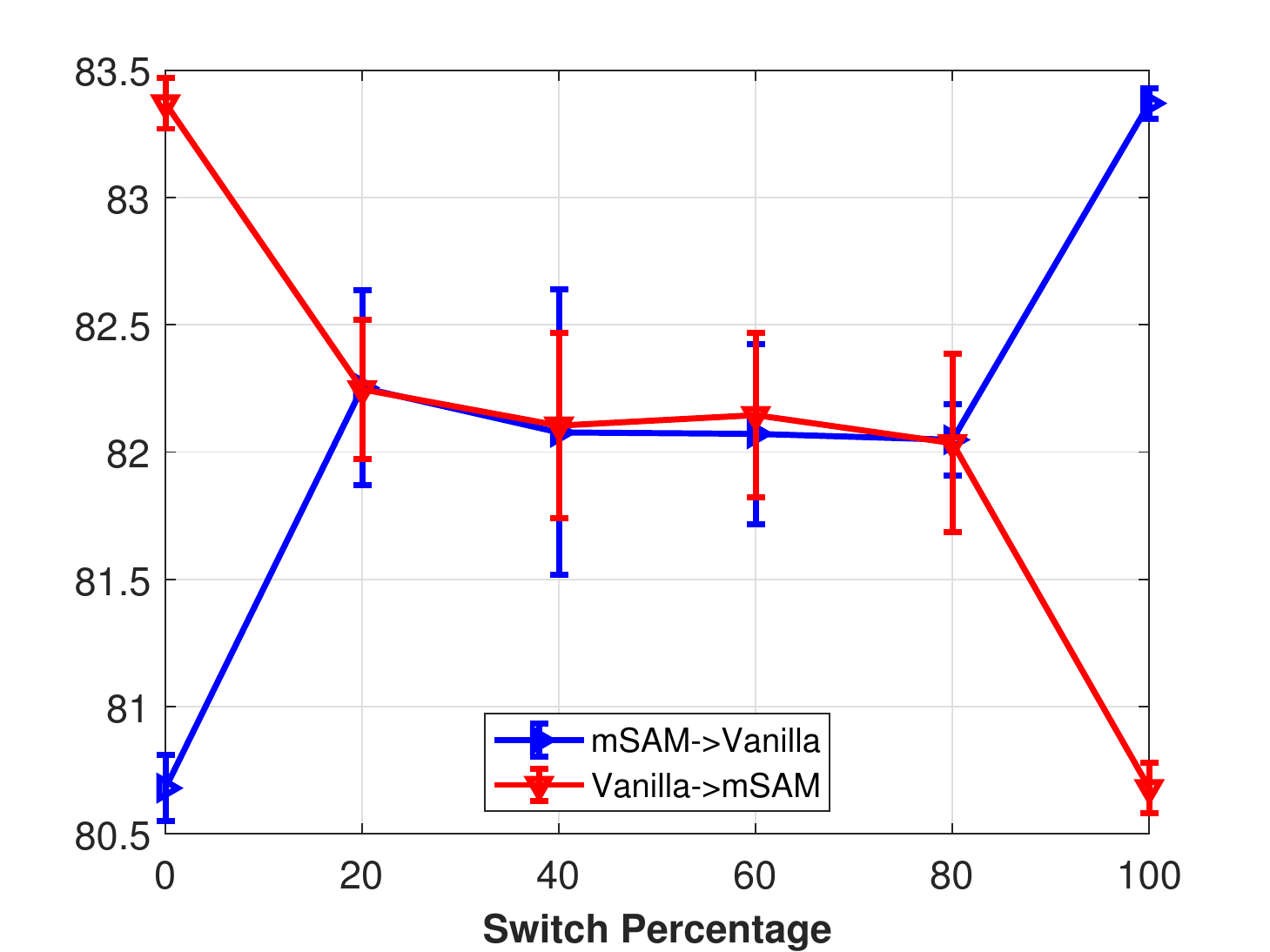}&   
     \includegraphics[scale=0.50]{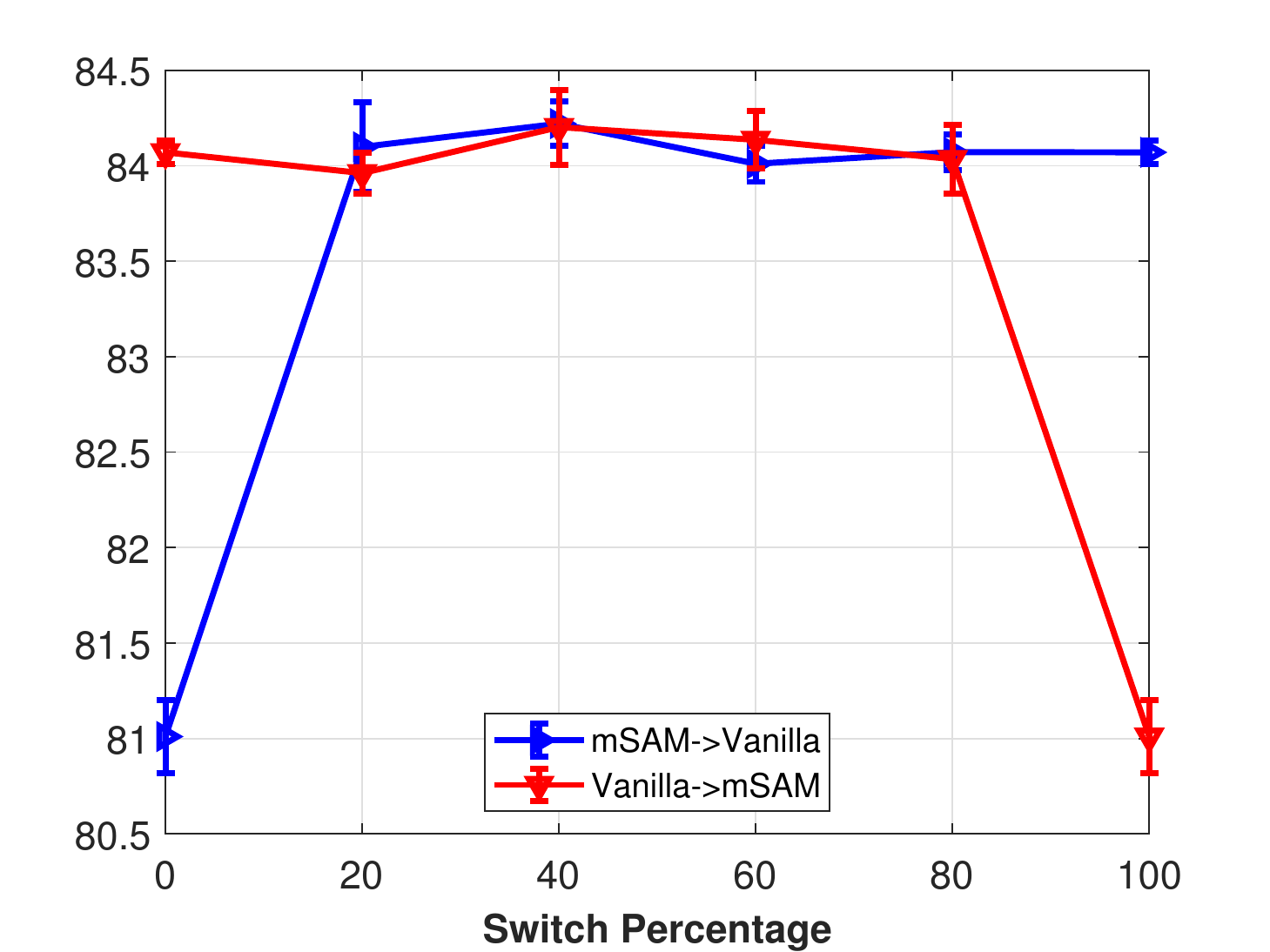}  
\end{tabular}
\label{fig:switch}
\end{figure}

Although mSAM does not appear to be computationally prohibitive in our experiments, it is still not as efficient as vanilla training, leaving room for further improving its efficiency. To that end, we conduct the following set of experiments. Building on our CIFAR100 experiments from Section~\ref{imagesec}, we start the training either with mSAM or vanilla training and then switch to the other training algorithm at some point. We keep all other training parameters fixed. The accuracy results for this setup for ResNet50 and WRN-28-10 are reported in Figure~\ref{fig:switch}. In this figure, the switch percent is the threshold in training when we transition from one algorithm to the other. For example, for the switch percent of $20$, if we start with mSAM, we use mSAM for the first $20\%$ of epochs and vanilla updates for the rest. If mSAM is used for the initial and/or final part of training, the accuracy is always better than vanilla training. In fact, in the WRN-28-10 case, as long as we partially use mSAM, the accuracy is almost the same as training with mSAM for the entire duration. For ResNet50, not using mSAM for the whole training leads to a drop in performance; however, even in this case, the accuracy of the hybrid training is better than the SAM training. These observations suggest that it is possible to enjoy the superior performance of mSAM, at least to some degree, while not having to deal with the computational complexity of mSAM for the entire training. A better theoretical and empirical understanding of the hybrid training method can be an exciting avenue for future work. 
%


\clearpage
\begin{table}[t]
\caption{Runtime of different methods and architectures on CIFAR100 data}
\label{runtime-table}
\begin{center}
\begin{tabular}{ccccc}
 Model & Vanilla & SAM &  mSAM 
\\  
\hline 
ResNet50      & $4497\pm11$ & $7440\pm9$  &  $16196\pm77$ \\
 WRN     & $10675\pm18$  & $17483\pm40$  &   $22261\pm24$\\
 ViT-B/16      & $4349\pm21$ & $7007\pm43$ &   $8163\pm14$ \\
\end{tabular}
\end{center}
\end{table}

\section{Hyper-parameters for Image Classification Experiments}\label{imageparam}
As mentioned, our experiments on CIFAR data in this section are done on 4 Nvidia V100 GPUs, with effective batch size of 512. For mSAM, we have used the micro-batch size of $16$ which corresponds to $m=32$. Rest of the hyper-parameters are chosen as in Table~\ref{cnn-params} for CNNs, and as in Table~\ref{vit-params} for ViT experiments.

\begin{table}[t]

\caption{Hyper-parameters for CNN experiments}
\label{cnn-params}
\begin{center}
\begin{tabular}{c|cc}
Model & ResNet50 & WRN-28-10 \\
\hline
Optimizer & \multicolumn{2}{c}{SGD}\\
Peak Learning Rate & 0.5 & 0.75 \\
Number of epochs &\multicolumn{2}{c}{200}\\
Momentum &\multicolumn{2}{c}{0.9}\\
Weight Decay &\multicolumn{2}{c}{$5\times 10^{-4}$}\\
Label Smoothing &\multicolumn{2}{c}{0.1}\\
Learning Rate Schedule &\multicolumn{2}{c}{One cycle with $5\%$ warm-up}\\
$\rho$ (SAM/mSAM) &\multicolumn{2}{c}{0.2}\\
\end{tabular}
\end{center}
\end{table}

\begin{table}[t]

\caption{Hyper-parameters for ViT experiments}
\label{vit-params}
\begin{center}
\begin{tabular}{c|c}
Model & ViT-B/16 \\
\hline
Optimizer & AdamW\\
Peak Learning Rate & $10^{-3}$ \\
Number of epochs &20\\
Weight Decay &0.3\\
Learning Rate Schedule &One cycle with $5\%$ warm-up\\
Gradient Clipping & norm=1\\
$\rho$ (SAM/mSAM) &0.3\\
\end{tabular}
\end{center}
\end{table}

\section{Hyper-parameters for GLUE Experiments}\label{glueparam}
Similarly, our experiments in this section are done on 4 Nvidia V100 GPUs, with effective batch size of 32. For mSAM, we have used the microbatch size of 4 which corresponds to $m=8$. The rest of task-specific hyper-parameters can be found in Table~\ref{nlp-params}.

\begin{table}[t]
\caption{Hyper-parameters for NLP experiments}
\label{nlp-params}
\begin{center}
\begin{tabular}{c|cccc}
Task & COLA & MRPC & SST-2 & QQP
\\ \hline 
Optimizer     & \multicolumn{4}{c}{AdamW}\\
Learning Rate       & $10^{-5}$ & $10^{-5}$  &$5\times 10^{-6}$ & $2\times 10^{-5}$\\
Learning Rate Schedule       & \multicolumn{4}{c}{One cycle with $6\%$ warm-up}\\
Number of Epochs        & 60 & 60  & 20 & 15\\
Weight Decay & \multicolumn{4}{c}{0.01}\\
$\rho$ (SAM/mSAM)      &0.01 & 0.01  & 0.05 & 0.05\\
\end{tabular}
\end{center}
\end{table}

\end{document}